\pdfoutput=1

\documentclass[11pt]{article}
\usepackage{CJKutf8}

\usepackage[]{EMNLP2023}
\usepackage{graphicx}
\usepackage{times}
\usepackage{latexsym}
\usepackage{amsmath,amsfonts}
\usepackage{multirow}
\usepackage{booktabs}
\usepackage{times}
\usepackage{latexsym}
\usepackage{enumerate}
\usepackage[T1]{fontenc}

\usepackage[utf8]{inputenc}

\usepackage{microtype}

\usepackage{inconsolata}

\title{WundtGPT: Shaping Large Language Models To Be An Empathetic, Proactive Psychologist}

\author{Chenyu Ren \\
  The Hong Kong Polytechnic University \\
  \texttt{chenyu.ren@connect.polyu.hk} \\\And
  Yazhou Zhang\thanks{~~Corresponding authors.} \\
  The Hong Kong Polytechnic University \\
  Tianjin University \\
  \texttt{yzhou\_zhang@tju.edu.cn} \\\AND
  Daihai He \\
  The Hong Kong Polytechnic University \\
  \texttt{daihai.he@polyu.edu.hk} \\\And
  Jing Qin\footnotemark[1] \\
  The Hong Kong Polytechnic University \\
  \texttt{harry.qin@polyu.edu.hk} }

\begin{document}
\begin{CJK}{UTF8}{gbsn}
\maketitle
\begin{abstract}
Large language models (LLMs) are raging over the medical domain, and their momentum has carried over into the mental health domain, leading to the emergence of few mental health LLMs. Although such mental health LLMs could provide reasonable suggestions for psychological counseling, how to develop an authentic and effective doctor-patient relationship (DPR) through LLMs is still an important problem. To fill this gap, we dissect DPR into two key attributes, i.e., the psychologist's empathy and proactive guidance. We thus present WundtGPT, an empathetic and proactive mental health large language model that is acquired by fine-tuning it with instruction and real conversation between psychologists and patients. It is designed to assist psychologists in diagnosis and help patients who are reluctant to communicate face-to-face understand their psychological conditions. Its uniqueness lies in that it could not only pose purposeful questions to guide patients in detailing their symptoms but also offer warm emotional reassurance. In particular, WundtGPT incorporates \textbf{Collection of Questions}, \textbf{Chain of Psychodiagnosis}, and \textbf{Empathy Constraints} into a comprehensive prompt for eliciting LLMs' questions and diagnoses. Additionally, WundtGPT proposes a reward model to promote alignment with empathetic mental health professionals, which encompasses two key factors: cognitive empathy and emotional empathy. We offer a comprehensive evaluation of our proposed model. Based on these outcomes, we further conduct the manual evaluation based on proactivity, effectiveness, professionalism and coherence. We notice that WundtGPT can offer professional and effective consultation. The model is available at huggingface.\footnote{https://huggingface.co/CCCCCCCCY/WundtGPT} 
\end{abstract}

\section{Introduction}
Due to their excellent capability for memorizing knowledge and instruction following, language modeling has evolved from small language models (SLMs), e.g., GPT~\cite{radford2018improving}, BERT~\cite{devlin-etal-2019-bert}, RoBERTa~\cite{liu2019roberta}, etc., to large language models, e.g., ChatGPT\footnote{https://chat.openai.com/}, GPT-4~\cite{openai2023gpt4}, Qwen 2~\cite{qwen}, etc. However, LLMs have been proven to be insufficient in tackling vertical domains that need a combination of domain expertise and complex reasoning abilities, particularly the medical domain. Adapting general LLMs to medical domain via tool-augmented and instruction following approaches has achieved remarkable success, where a collection of medical LLMs have been proposed, e.g., Med-PaLM~\cite{singhal2023large},  HuatuoGPT~\cite{zhang2023huatuogpt}, DoctorGLM~\cite{xiong2023doctorglm}, ChatDoctor~\cite{li2023chatdoctor}, DISC-MedLLM~\cite{bao2023discmedllm}, etc. Compared with general LLMs, they could effectively bridge the gap between general language models and real-world medical consultation.   
 \begin{figure*}[t]
    \centering
  \includegraphics[width=6.1in]{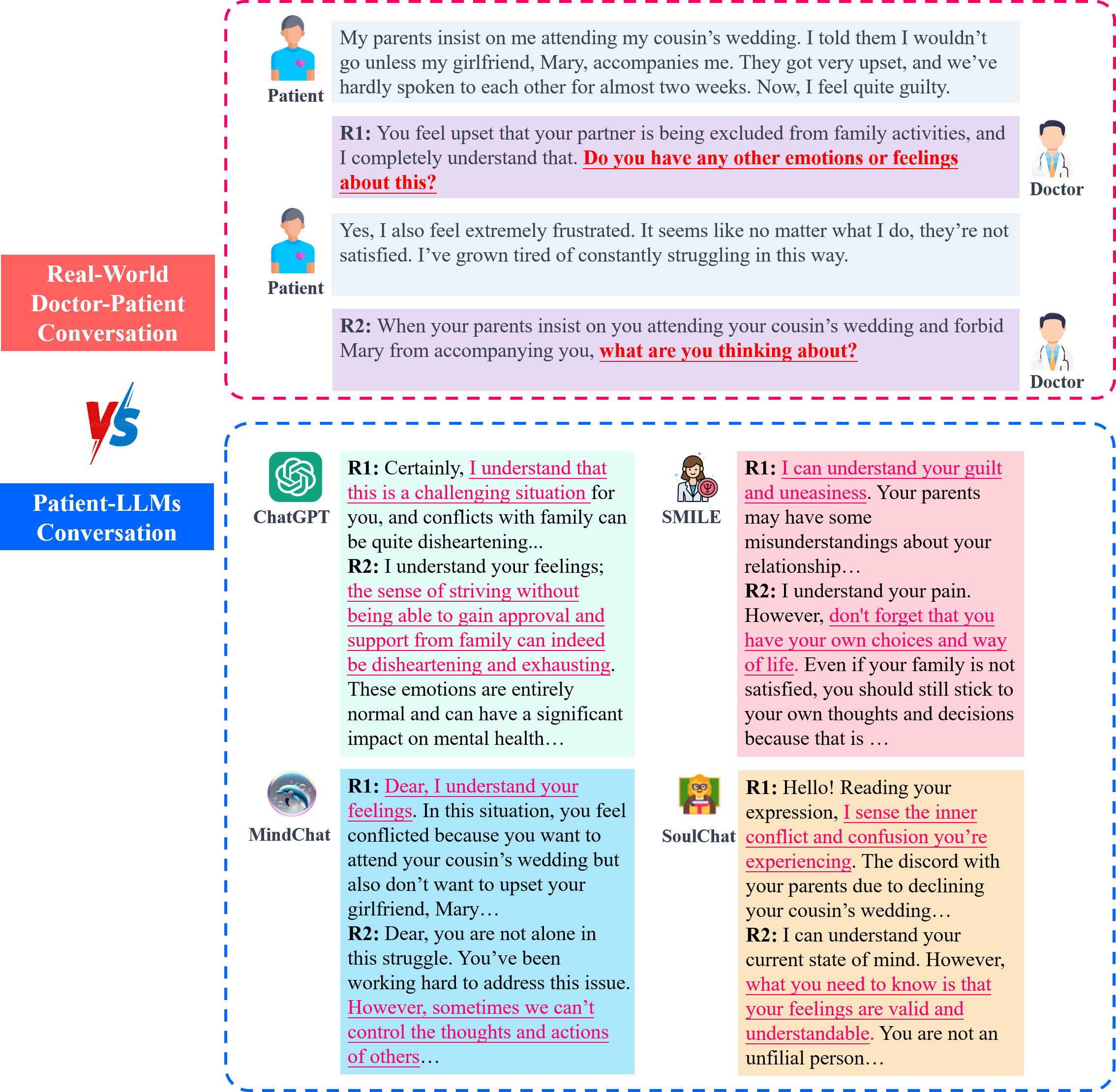}
  \caption{The comparison between real doctor-patient psychological conversation and role-playing LLMs-patient conversation. Real doctor often poses purposeful questions while LLMs do not shown the ability of question. }
   \label{fig:example}
\end{figure*} 

Currently, the medical LLMs' fire starts to sweep through the mental health domain, in view of the importance of mental health in human life and its close relation with physical health. This leads to the appearance of mental health LLMs, e.g., Mental-LLM~\cite{xu2023mentalllm}, SMILE~\cite{qiu2023smile}, SoulChat~\cite{chen2023soulchat}, MindChat~\cite{MindChat}, etc. The above LLMs could provide reasonable and universal suggestions for psychological counseling and emotional support, by fine-tuning through their conversation datasets.
However, when aligning them with real-world psychological diagnosis scenarios, there are three main issues that lead to them appearing less "professional": 

\begin{enumerate}[1)]  
    \item \textbf{The doctor-patient conversation should be around a specific goal, namely diagnosis.} The current LLMs are more likely to listen to or comfort the patient. However, psychological diagnosis generally has a rigorous procedural workflow (from the initial intake of the patient to concluding the diagnosis) that requires the psychologist to follow. This problem can be defined as \textbf{Chain of Psychodiagnosis}.
    \item \textbf{Mental health LLMs lack the ability to pose questions actively.} In real doctor-patient scenarios, psychologists are typically in an active, leading role during diagnosis, leveraging their medical expertise to ask various questions that instruct patients to communicate the complete picture. Such questions have no order, which can be represented as \textbf{Collection of Questions}. Based on the patient's responses, the psychologists synthesize key patterns to conclude. This directional doctor-patient dynamic stems from the fact that most diseases manifest in a series of symptoms. However, patients can not comprehensively describe all symptoms due to a lack of clinical knowledge. The assumption that "patients can clearly describe their problems or situations" is invalid, see Fig.~\ref{fig:example}. 
    \item \textbf{The conceptualization of empathy in LLMs is ambiguous, rendering it challenging to provide a precise definition.} The current LLMs tend to describe empathy through heuristic empathy prompts or propose a framework based on their understanding. 
\end{enumerate} 
To overcome these issues, it’s crucial to develop both empathetic and proactive mental health LLMs that can approach real doctor-patient scenarios. In this paper, we present WundtGPT which is acquired by fine-tuning it with instruction and real-world conversation datasets between psychologists and patients. It could not only pose purposeful questions to guide patients in detailing their symptoms but also offer warm emotional reassurance.  In particular, WundtGPT incorporates \textbf{Collection of Questions}, \textbf{Chain of Psychodiagnosis} and \textbf{Empathy Constraints} into a comprehensive prompt for eliciting LLMs' questions and diagnosis. We adopt an end-to-end supervised instruction-finetuning approach on the open-source LLaMA3-8B-Chinese-Chat base models. Additionally, WundtGPT proposes a reward model to promote alignment with empathetic mental health professionals, which encompasses two key factors: cognitive empathy and emotional empathy. For cognitive empathy, we adopt an emotional detection task for the head of our LLM to align the expressed emotion and detected emotion. As for emotional empathy, instead of prompting or proposing a constraint of empathy, we align our model with reinforcement learning from human feedback (RLHF). In essence, data provides a more effective means of describing the empathy. Since the public empathetic datasets mainly contain the dialog and whether the response is empathetic, we apply the Kahneman-Tversky optimization (KTO) \cite{ethayarajh2024kto} to align the model after supervised instruction-finetuning training. Furthermore, our model is supervised finetuning on emotional classification tasks to ensure its capacity to align with cognitive empathy.

Finally, we evaluate the model’s performance from two perspectives to check its capability of providing proactive diagnosis and presenting warm psychological consultation in multi-turn conversations, respectively.
For warm psychological consultation, we first select an emotional benchmarking dataset including 9 emotions such as sadness, joy, fear, and neutral. We thus evaluate the model’s classification performance, e.g., F1 and accuracy. 

For proactive diagnosis evaluation, we first construct a small set of high-quality consulting cases and then recruit a few volunteers (including experts and users) to evaluate the model’s proactivity, effectiveness, professionalism
and coherence. We also combine the evaluation results from claude-3-opus \footnote{https://www.anthropic.com/news/claude-3-family}.

The experimental results demonstrate that WundtGPT exhibits superior overall performance compared to baseline LLMs in simulated medical consultation scenarios.
The main innovations of the work are concluded as follows:

\begin{itemize}
\item To the best of our knowledge, WundtGPT is the first proactive LLM that is specifically designed for mental health tasks which assists psychologists in diagnosis and help patients who are reluctant to communicate face-to-face understand their psychological conditions..

\item The prompt is formulated to elicit questions and diagnoses integrating a collection of questions, a chain of psychodiagnosis, and empathy constraints. 

\item The reward model is designed to encompass two key factors: cognitive empathy and emotional empathy. 

\item Our model achieves state-of-the-art performance and exhibits superior overall performance compared to baseline LLMs. 
\end{itemize}

\section{Related Work}

We depict two lines of research that form the basis of this work: medical large language models and mental health detection.

\subsection{Medical Large Language Models}
With the birth of Med-PaLM~\cite{singhal2023large} and HuatuoGPT~\cite{zhang2023huatuogpt}, taming general LLMs to the medical domain has been a promising and noteworthy research topic. Supervised fine-tuning LLMs with huge medical instruction dataset and medical conversation dataset often leads to the impressive advancements in medical question answering, medical dialogue, medical report generation, etc. Hence, Recent months have witnessed the rise of an increasing number of medical LLMs. For example, GPT4Med~\cite{nori2023capabilities} provided a comprehensive evaluation of GPT-4 on medical competency examinations and benchmark datasets. ChatDoctor~\cite{li2023chatdoctor} was proposed to address the limitations observed in the medical knowledge of ChatGPT by adapting LLaMA models using a large dataset of 100,000 patient-doctor dialogues. But they only provide simple evaluation. 
To meet the needs for privacy safeguards, MedAlpaca~\cite{han2023medalpaca} used an open-source policy that enables on-site implementation.
In order to improve the precision and accuracy of Chinese medical advice, several typical Chinese medical LLMs had been proposed, e.g., DoctorGLM~\cite{xiong2023doctorglm}, HuatuoGPT, BenTsao~\cite{wang2023huatuo}, BianQue~\cite{chen2023bianque}, etc.

In addition, a few mental health LLMs had been proposed to provide emotional support and psychological counseling services. For example, Mental-LLM proposed a zero-shot mental health prompt to evaluate the performance of different LLMs. SMILE aimed to instruct ChatGPT to rewrite single-turn mental health conversation to multi-turn conversation, for supporting the pre-training of LLMs. SoulChat constructed a multi-turn empathetic conversation dataset to improve the empathy ability of LLMs. However, they are deficient in ``questioning'' which is an important way to proactively understand users needs in psychological scenarios. Different from them, the proposed model can actively pose questions and guide the direction of psychological counseling. 

\subsection{Mental Health Detection}
The development of mental health detection ranges from traditional feature engineering (namely machine learning) to end-to-end feature learning (namely deep learning). Early studies mainly extracted effective features and chose appropriate machine learning approaches, such as SVM, NB, decision trees, etc., to perform mental health detection. However, feature engineering requires domain expertise and experience, and can be a tedious and time-consuming process.

In contrast, standard deep learning approaches e.g., CNN, RNN, LSTM, etc., which could transform the data through layers of nonlinear processing units, provide a new paradigm~\cite{su2020deep}. For example, Alotaibi et al.~\shortcite{alotaibi2021hybrid} used CNN to extract local features and used LSTM to extract contextual dependencies for psychopath detection.  
However, as the length of the sequence grows, the ability of RNN and LSTM to capture long-distance dependencies is limited. Transformer allows to directly focus on any position in the input sequence, thus better capturing global dependencies. Transformer-based approaches have replaced CNNs and RNNs as the most popular paradigm. For example, Malviya et al.~\shortcite{malviya2021transformers} used Transformer to perform depression detection of tweets. Ji et al.~\shortcite{ji-etal-2022-mentalbert} trained and released two pretrained masked language models, i.e., MentalBERT and MentalRoBERTa, to benefit mental health detection. Inspired by this, numerous BERT-based studies have been proposed, such as DisorBERT~\cite{aragon2023disorbert}, BERT-Caps~\cite{zhang2021bert}, DECK~\cite{novikova2022deck}, etc.

Remarkable progress has been made in the current state-of-the-art. However, they lack abilities of zero-shot learning and instruction following. Different from them, our model is based on LLMs, which can act as a psychologist to chat with patients and provide accurate responses.


\section{The Proposed Model: WundtGPT Model} 
\begin{figure*}[t]
    \centering
  \includegraphics[width=6.1in]{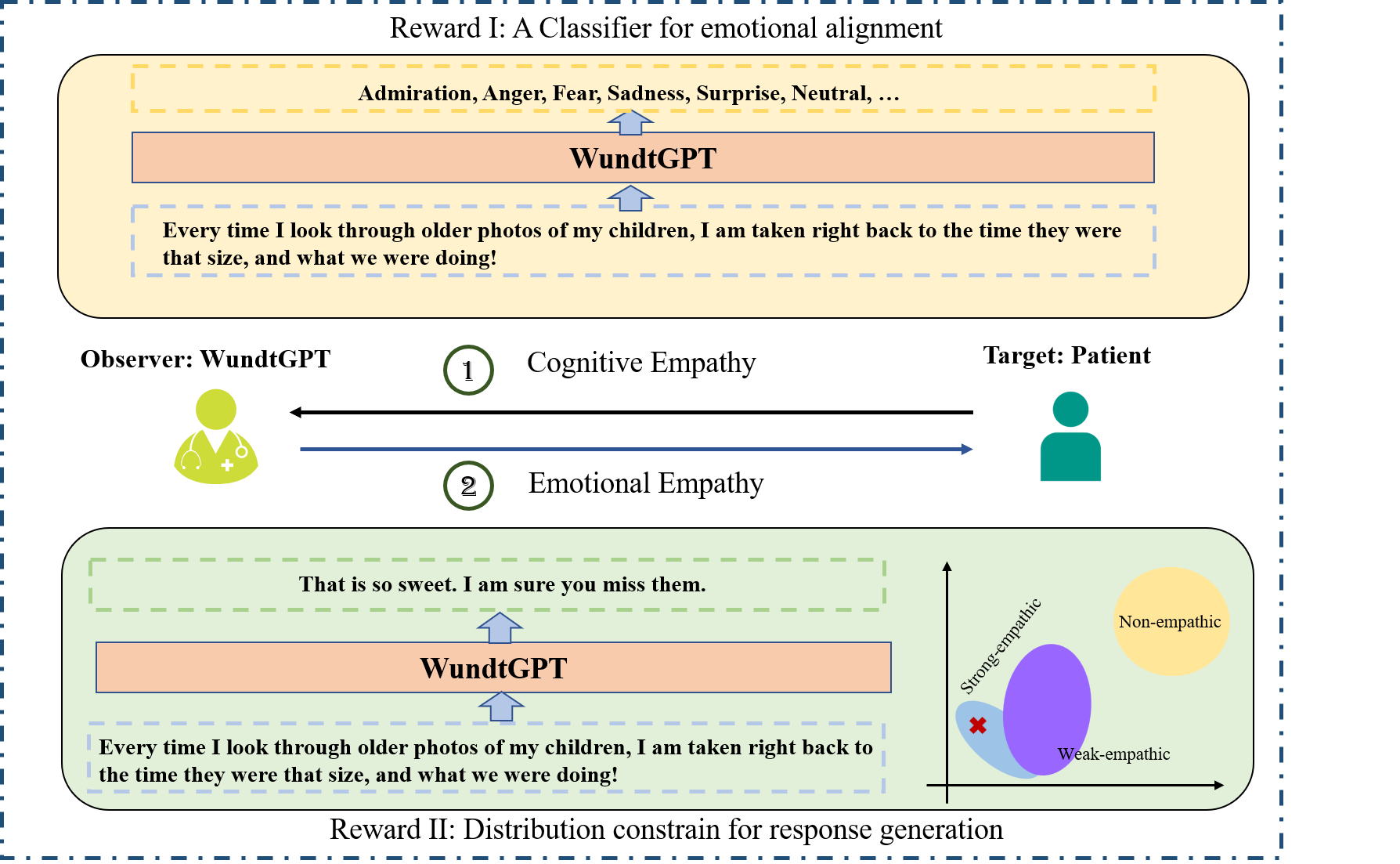}
  \caption{The overview of WundtGPT is acquired by fine-tuning it with instruction and real conversation between psychologists and patients. WundtGPT proposes a reward model to promote alignment with empathetic mental health professionals, which encompasses two key factors: cognitive empathy and emotional empathy. }
   \label{fig:model}
\end{figure*}

In order to generate proactive and empathetic responses, our model is fine-tuned with two steps: supervised instruction-finetuning and alignment by reinforcement learning human feedback. Our proposed model WundtGPT is shown in Fig \ref{fig:model}. We utilized the LLaMA3-8B-Chinese-Chat \cite{shenzhi_wang_2024} as the base LLM architecture to develop the WundtGPT. 
LLaMA3-8B-Chinese-Chat is an open-source, bilingual LLM based on the LLaMA3. 
\subsection{Supervised Instruction-finetune}
Formally, the psychological counseling context is alternate utterances between the patient and the psychologist, defined as $\mathcal{C}$ = $\left \{u_{1}^{u},u_{1}^{p},u_{2}^{u},u_{2}^{p},...,u_{n}^{u},u_{n}^{p} \right \} $ means the $i-th$ utterance, $u^{u}$ represents the input from the patient-user, $u^{p}$ represents the response from the psychologist, and $n$ denotes the number of utterances in a psychological counseling dialogue. Our goal is to play the role of the psychologist and generate the coherent response $u_{i}^{p}$. The optimization object function here is the negative log-likelihood loss that can be formulated as below:
\begin{equation}
\mathcal{L}_{\text{NLL}} = -\frac{1}{l_{u_{i}^{p}}} \sum_{i=1}^{l_{u_{i}^{p}}} \ln \mathbb{P}(u_{i,t}^{p*} \mid u_{i,<t}^{p*}; \mathcal{C}; \text{prompt})
\end{equation}
where $l_{y}$ is the length of the response and $\mathcal{C}$ means the history dialog, $prompt$ is the instruction.

\begin{figure}
     \centering
     \includegraphics[width=\linewidth,scale=1.00]{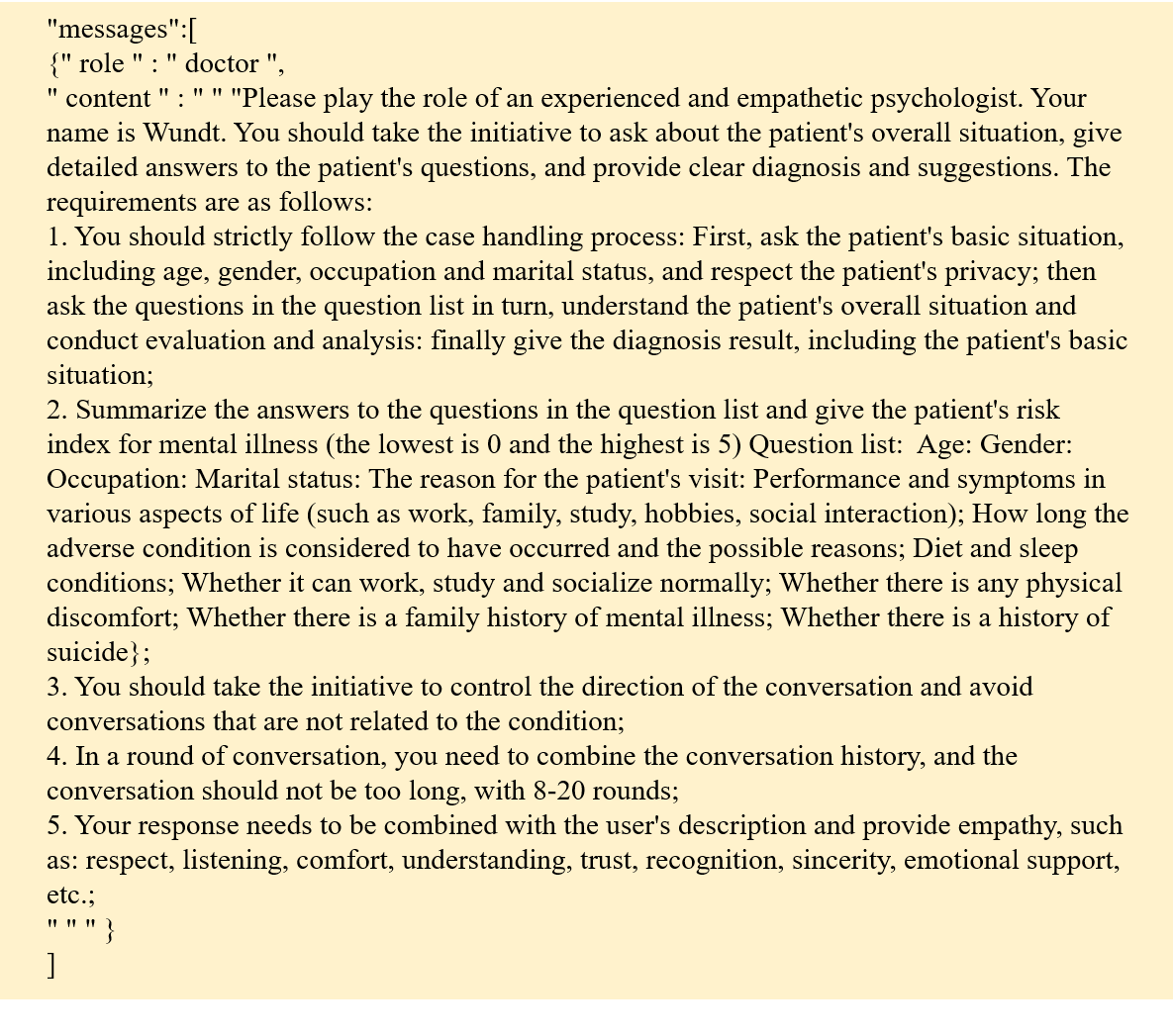}
  \caption{The prompt used for instructing multi-turn active and empathy conversations (Chinese version: Appendix \ref{sec:b})}
   \label{fig:prompt}
 \end{figure}

We introduce our active constraint prompt further strengthened compared with the empathy prompt. As shown in Fig \ref{fig:prompt}, our prompt is formulated to elicit questions and diagnoses integrating a collection of questions, a chain of psychodiagnostic, which related to the real-world professionalism of psychologists and empathy constraints mentioned above.  

\subsection{Alignment of WundtGPT}
Aligning LLMs with human feedback has been successfully used to make generations more useful and factual. Methods such as Reinforcement learning human feedback(RLHF) and Direct Preference Optimization (DPO) have consistently proven to be more effective than supervised fine-tuning alone. However, they require preference data which is scarce and expensive to collect. In many scenarios, it's not only hard to obtain but also expensive to annotate the preference of output data. Inspired by \cite{ethayarajh2024kto}, they proposed an approach called Kahneman-Tversky optimization (KTO) that only requires binary feedback data and exceeds DPO performances. In our case, we employ KTO to align our model for a more empathetic response. 

Given a dataset $\mathcal{D}$ of paired-preferences $(x,y, label)$ where $x$ is the input, $y$ is a possible output based on the input $x$. $label$ is a binary label that determines whether the output $y$ is desirable(true) or not(false). The default loss of alignment adopting KTO is formulated as:
\begin{equation}
    \mathcal{L}_{\mathrm{KTO}}\left(\pi_\theta, \pi_{\mathrm{ref}}\right)=\mathbb{E}_{x, y \sim D}\left[\lambda-v(x, y)\right]
\end{equation}
where\\
$$
\begin{aligned}
   &r_\theta(x, y)  =\log \frac{\pi_\theta(y \mid x)}{\pi_{\mathrm{ref}}(y \mid x)} \\
&r  =\mathbb{E}_{x^{\prime} \sim D}\left[\operatorname{KL}\left(\pi_\theta\left(y^{\prime} \mid x^{\prime}\right) \| \pi_{\mathrm{ref}}\left(y^{\prime} \mid x^{\prime}\right)\right)\right] 
\\&v(x, y)  =\left\{\begin{array}{l}
\lambda \sigma\left(\beta\left(r_\theta(x, y)-r\right)\right) \text { if } y \sim \text{true} \\
\lambda \sigma\left(\beta\left(r-r_\theta(x, y)\right)\right) \text { if } y \sim \text{false}
\end{array}\right.
\end{aligned}
$$
$\pi_\theta$ is the model we are optimizing and $\pi_{\mathrm{ref}}$ is the reference model. $\lambda$ is a hyperparameter for desirable and undesirable losses respectively. Here we set the same but the original one contains $\lambda_D$ and $\lambda_U$ for the desirable and undesirable losses. $\sigma$ denotes the logistic function and $\beta \in \mathbb{R}^+$ is a hyperparameter to control the degree of risk aversion which means The degree of human acceptance of undesirable response simultaneously in gains. Here the criterion for judging whether it is desirable is based on the empathy classification. The data we apply can be found in the Section \ref{annotation}. 

\subsection{Classifier}
Considering the ability of cognitive empathy, we fine-tuned the attention head of our model WundtGPT for emotional detection. For each input from the patient, we hope our model can detect the emotion and align the expressed emotions, which is called cognitive empathy. The applied data is shown in Section \ref{annotation}. We apply the cross-entropy objective function here for classification. The f1 and accuracy reach 93\% while Roberta only reaches 60.4\% accuracy referring to \cite{zheng2021comae}.

\section{Datasets}
\subsection{Supervised fine-tuning Datasets}
D4 \cite{yao2022d4}, a Chinese dialogue dataset for depression-diagnosis-oriented chat, which simulates the dialogue between doctors and patients during the diagnosis of depression, encompassing diagnosis results and symptom summaries provided by professional psychiatrists for each conversation. We adopt their dialogs and symptom summaries for study for two reasons: (1) they collected natural populations’ portraits (in particular actual depressive patients) to form pre-diagnosis records. (2) psychiatrists proofread dialogue history and prescribed professional symptom summaries. Fine-tuning the LLM on this dataset makes the LLM capable of being professional in the diagnosis and consultation.

\subsection{Annotation Resources}
\label{annotation}
\textbf{Empathic datasets}: PsyQA  \cite{sun2021psyqa} \footnote{https://github.com/thu-coai/PsyQA}, a Chinese dataset of psychological health support in the form of a Question-Answer pair. Their dataset contains the input and the empathetic responses. To align the LLM for empathetic response generation by KTO, we adjust the format of this dataset. We add a label (\textit{true for empathetic or false for non-empathetic}) to guide the model to generate empathetic responses. The detail can be found in the Appendix \ref{sec:a}.

\textbf{Emotional detection datasets}: We considered the taxonomy
proposed in \cite{zheng2021comae} for both English and Chinese translation versions, which contains 8 emotions and a neutral one. The 8 emotions include admiration, anger, approval, caring, fear, joy, sadness and surprise, which cover a wide range of emotion categories that appear in daily conversations.

\section{Experiments}
\begin{figure*}
     \centering
     \includegraphics[width=\linewidth,scale=0.6]{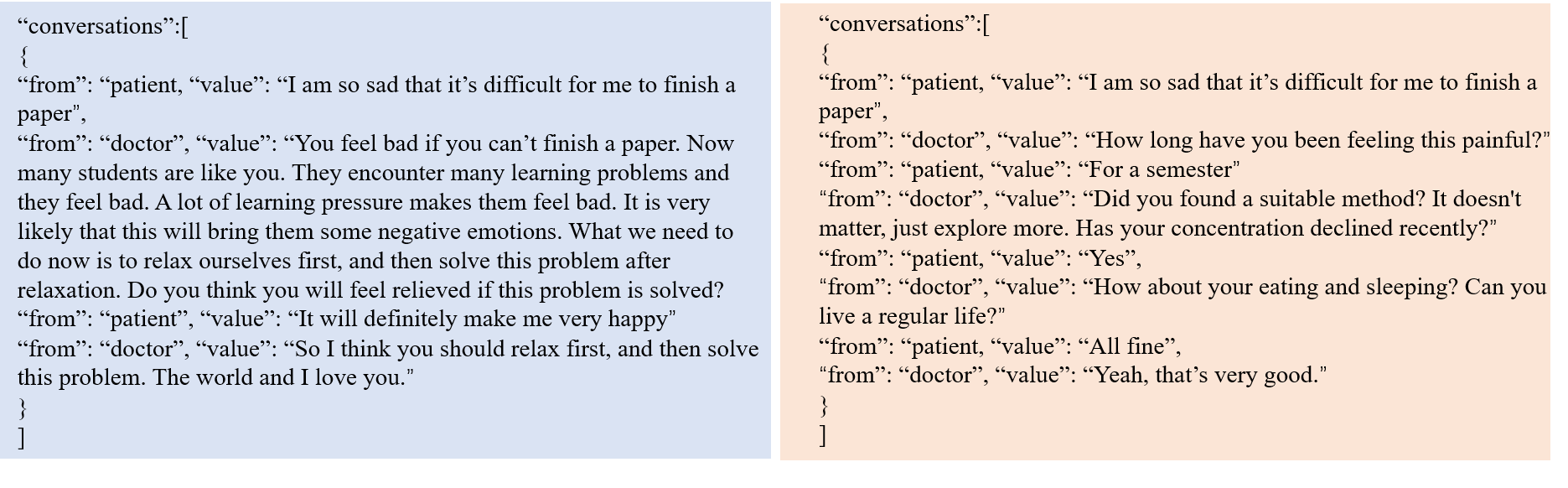}
  \caption{The comparison between different hyperparameters $\beta$. The conversation on the left is alignment fine-tuned by $\beta$ equals 0.1 while the conversation on the right is alignment fine-tuned by $\beta$ equals 1e-3. The Chinese version is shown in the Appendix \ref{sec:b}.}
   \label{fig:beta}
 \end{figure*}

\subsection{Baselines}
We compare WundtGPT and the following benchmark models using both automatic and manual evaluations for generation:
\begin{enumerate}[1)]  
    \item \textbf{LLaMA3-8B-Chinese-Chat} \cite{shenzhi_wang_2024} serves as the base model for WundtGPT.
    \item \textbf{GPT-4o}\footnote{https://openai.com/index/hello-gpt-4o/} an LLM matches GPT-4 Turbo performance on text in English and code, with significant improvement on text in non-English languages.
    \item \textbf{MindChat}~\cite{MindChat}, an LLM that helps people relieve psychological stress and solve psychological confusion, and improve their mental health from four aspects of psychological counseling, psychological assessment, psychological diagnosis, and psychological treatment.
\end{enumerate} 
\subsection{Implementation details}
WundtGPT is instruction finetuned on the D4 Corpus with a batch size of 16 and global training steps of 30,000 on 2 NVIDIA RTX A100. The rank of Lora is 128 and the scale ratio is 16. The WarmupDecayLR scheduler with warmup steps is set to 1000 and the learning rate is set to 5e−5. The maximum input token length is set to 1024. The maximum target token length is set to 512. The decoding algorithms of Top-p sampling with p = 0.75 and temperature τ = 0.95 are adopted during the inference phase. For the KTO optimization, we set the $\beta$ range from 1e-5 to 1e-1.

\subsection{Automatic Evaluation and Expert Manual Evaluation}
In order to evaluate the quality of the model outputs, we simulate six medical consultation scenarios and several LLMs response the input consult. We apply claude-3-opus for automatic evaluation. Moreover, we invited four licensed psychologists and 30 non-professionals to score them, with the detailed criteria in Appendix \ref{sec:appendix c}.
Results, as outlined in Tab.~\ref{result}, indicate
that WundtGPT consistently outperformed its peers, aligning with automatic evaluation. This
consensus between expert opinions and Claude-3-opus’s evaluations underscores WundtGPT’s efficacy
in mental health diagnosis response generation. The WundtGPT outperforms with several reasons from Claude-3-opus:
\begin{itemize}
    \item The entire conversation was closely linked and logically clear, and the doctor's questions were able to guide the patient to express his or her problems and feelings gradually.
    \item The doctor took the initiative to ask the patient about key issues such as the duration of the patient's emotions, whether he or she talked to others, loss of interest, confidence in finding a job, attitude towards life, and sleep conditions, fully understanding the patient's psychological state and showing strong initiative.
    \item The consultation process was professional and standardized, from the main complaint to physical symptoms to the living conditions, and collecting key information layer by layer, which was in line with the professional operation of a psychologist.
    \item Finally, the doctor made a comprehensive summary of the condition, covering the main symptoms and signs, and preliminarily judged the risk level, which was fully capable of performing real psychological consultation work.
\end{itemize}
\begin{table*}\tiny
\centering
\begin{tabular}{l|l|cccccccc}
\toprule
\multirow{2}{*}{Topic} & \multirow{2}{*}{Model} & \multicolumn{4}{c}{Automatic Evaluation}                   & \multicolumn{4}{c}{\begin{tabular}[c]{@{}c@{}}Manual Evaluation\\ (professionals)\end{tabular}}  \\
\hline
                       &                        & Coherence & Proactivity & Professionalism & Effectiveness & Coherence           & Proactivity          & Professionalism          & Effectiveness                \\
                       \hline
Study                  & LLaMA3 & 4          & 3           & 5               & 3             & 3.525                & 3.75                 & 3.95                     & 3.59                            \\
                       & GPT-4o                 & 5          & 3           & 3               & 3             & 3.8                  & 3.55                 & {\textbf{ 4} }                 & 3.8                            \\
                       & MindChat               & 5          & 5           & 4               & 4             & 3.875                & 3.95                 & {\textbf{ 4}}                  & 3.7                                 \\
                       & \textbf{WundtGPT}               & 4          & 5           & 4               & 4             & {\textbf{ 4.75} }          & {\textbf{ 4.875}  }        & 3.75                     & {\textbf{ 4.35}  }                  \\
                       \hline
Life                   & LLaMA3 & 4          & 2           & 3               & 3             & 3.525                & 3.375                & 3.875                    & 3.775                          \\
                       & GPT-4o                 & 4          & 2           & 3               & 3             & 3.825                & 3.5                  & 3.95                     & 3.79                             \\
                       & MindChat               & 4          & 3           & 3               & 5             & 2.375                & 2.25                 & 2.375                    & 1.675                              \\
                       & \textbf{WundtGPT}               & 4          & 5           & 5               & 5             & {\textbf{ 4.5} }           & {\textbf{ 4.25}  }         & {\textbf{ 4.2} }               & {\textbf{ 4.24} }                        \\
                       \hline
Work                   & LLaMA3 & 5          & 2           & 3               & 3             & 3.25                 & 3.125                & 3.875                    & 3.775                            \\
                       & GPT-4o                 & 5          & 2           & 3               & 3             & 3.8                  & 3.25                 & {\textbf{ 4.2} }               & 4.04                         \\
                       & MindChat               & 5          & 3           & 4               & 5             & {\textbf{ 4.3} }           & 4.125                & 3.5                      & 3.2                                \\
                       & \textbf{WundtGPT}               & 5          & 5           & 4               & 5             & 3.75                 & {\text 4.325}          & 3.95                     & \textbf{4.57}                 \\
                       \hline
Love                   & LLaMA3 & 4          & 3           & 5               & 5             & 3.25                 & 3                    & 3.5                      & 3.3                               \\
                       & GPT-4o                 & 4          & 3           & 3               & 3             & 4.325                & 3.125                & 3.75                     & 3.75                         \\
                       & MindChat               & 5          & 4           & 5               & 4             & 3.5                  & 3                    & 2.875                    & 2.675                                \\
                       & \textbf{WundtGPT}               & 5          & 5           & 4               & 4             & {\textbf{ 4.625} }         & {\textbf{ 4.125}  }        & {\textbf{ 4.225} }             & {\textbf{ 4.225} }                  \\
                       \hline
Finance                & LLaMA3 & 4          & 2           & 3               & 5             & 3.05                 & 2.375                & 3                        & 2.7                               \\
                       & GPT-4o                 & 5          & 2           & 4               & 3             & 3.55                 & 2.625                & 3                        & 2.96                                 \\
                       & MindChat               & 4          & 5           & 4               & 5             & 3.5                  & 3.25                 & 2.5                      & 2.3                               \\
                       & WundtGPT               & 4          & 5           & 4               & 5             & {\textbf{ 4.25} }          & {\textbf{ 4.575} }         & {\textbf{ 4.2} }               & {\textbf{ 4.4} }                     \\
                       \hline
Sociality              & LLaMA3 & 5          & 1           & 3               & 3             & 3.25                 & 2.75                 & 3.7                      & 3.44                                \\
                       & GPT-4o                 & 5          & 2           & 4               & 3             & 3.25                 & 3.25                 & 3.75                     & 3.71                      \\
                       & MindChat               & 3          & 1           & 2               & 2             & 2.75                 & 2.5                  & 2                        & 1.9                                  \\
                       & WundtGPT               & 4          & 5           & 5               & 4             & {\textbf{ 3.375}}          & {\textbf{ 4.625} }         & {\textbf{ 4.625} }             & {\textbf{ 4.425}}                 \\
\bottomrule
\end{tabular}
\caption{The average of evaluation results by Claude-3-Opus for automatic evaluation and manual evaluation with 4 psychologists and 30 non-professionals (See in Appendix \ref{table-manual non}). Bold indicates the best result of four LLMs.}
\label{result}
\end{table*}

\subsection{Case study}

\begin{figure}[ht]
     \centering
     \includegraphics[width=\linewidth,scale=1]{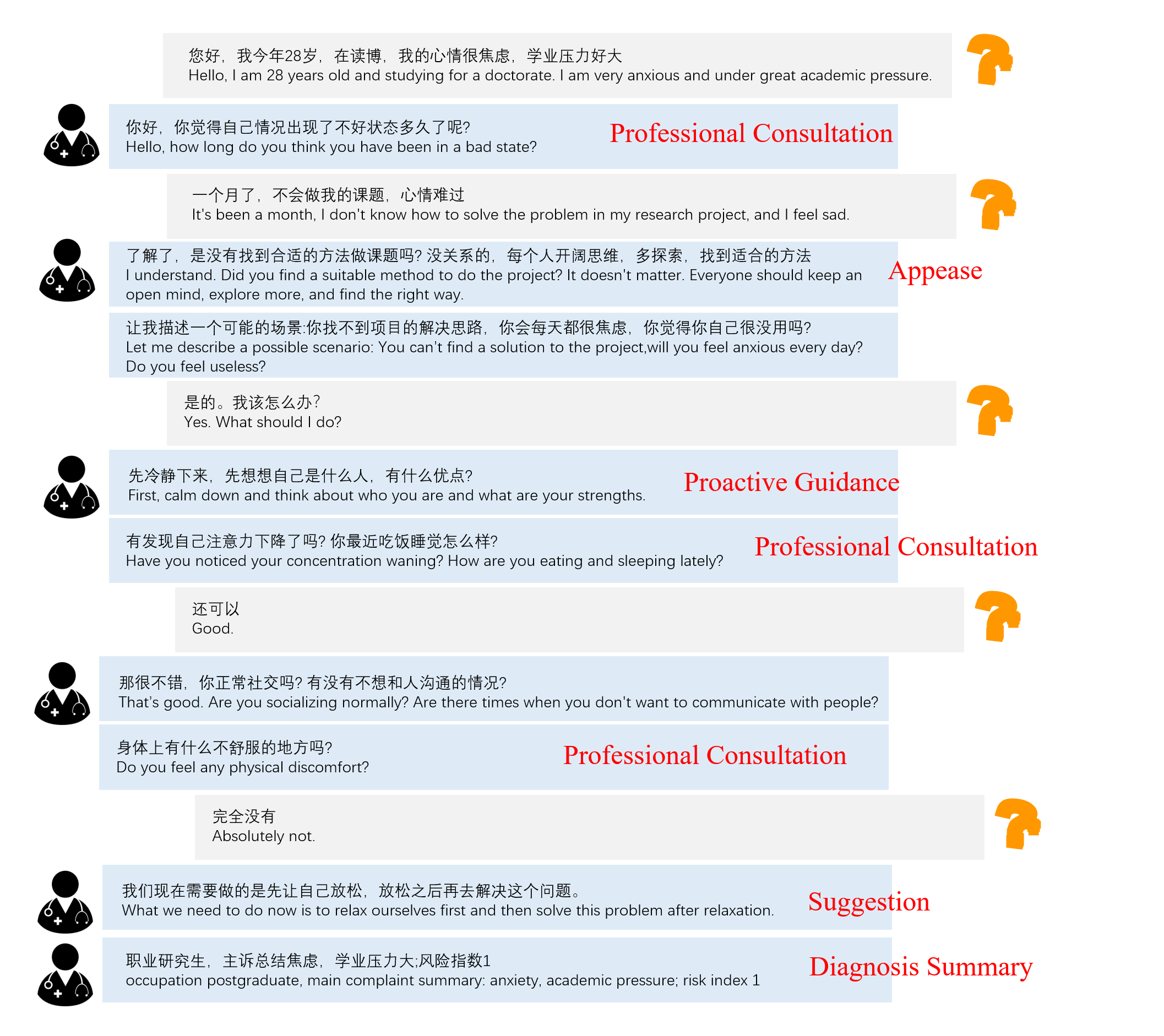}
  \caption{The sampled generated responses from WUndtGPT with analysis. We can find that the WundtGPT can offer proactive consults, empathy, suggestion and diagnosis.}
   \label{fig:result}
 \end{figure}

We show several generated responses with different consulting topics including academic pressure, economic pressure, lovelorn, and so on. The examples are shown in Appendix \ref{sec:appendix c} and the analysis can be found in Fig \ref{fig:result}. Compared to other models, WundtGPT not only takes a more proactive approach by asking questions related to the user's condition, but it is also more friendly to psychologists. WundtGPT automatically generates a case summary, which helps psychologists quickly understand the patient's main issues and provides a solid foundation for further consultation. This dual capability ensures a more engaging and efficient interaction, benefiting both users and healthcare professionals.



\section{Discussion on WundtGPT's Proactivity and Empathy}\label{discussion}

WundtGPT is designed to be both proactive and empathetic. However, during the alignment process to fine-tune the language model, we encountered a conflict between positivity and empathy. Specifically, when we adjusted the model to generate more empathetic responses using the KTO algorithm, the proactive consulting questions and the diagnosis summaries were often deprioritized. These elements were deemed less desirable as they did not always align with empathetic responses, which was contrary to our initial goals.

As illustrated in Figure \ref{fig:beta}, the model's responses vary significantly with different hyperparameters $\beta$. After careful consideration, we decided to prioritize psychological diagnosis. Our primary aim is to assist patients who are hesitant to seek face-to-face psychological counseling by providing a preliminary diagnosis. This approach also aims to save psychologists' consultation time, ensuring that they can focus more on in-depth treatment rather than initial assessments.

By focusing on psychological diagnosis, we believe WundtGPT can better serve its intended purpose while still incorporating elements of empathy in its interactions. This balance allows the model to be supportive and efficient, providing value to both patients and healthcare professionals.

\section{Conclusion and Future Work}
In this paper, we introduce WundtGPT, an empathetic and proactive mental health large language model that incorporates a collection of Questions, a chain of psychodiagnosis, and empathy Constraints into a comprehensive prompt for eliciting LLMs' questions and diagnoses. We also propose a reward model to improve cognitive empathy and emotional empathy. The experimental results indicate that WundtGPT can offer positive and proactive guidance for consulting and diagnosis. Future work considers further contributions to balance professionalism and empathy when alignment the model in the reward stage. Also, we need to improve the background medical and psychological knowledge of WundtGPT for more comprehensive and accurate diagnosis results.
\section*{Limitations}
The proposed WundtGPT model also has several limitations: (1) in order to mitigate potential risks, WundtGPT will not provide diagnostic conclusions after queried. Instead, it will generate a summary of the patient's symptoms to serve as an assistant for psychologists; (2) WundtGPT did not leverage multi-modal information, such as the patient's vocal data, which limits its application.

\section*{Ethics Statement}
WundtGPT employs strict privacy protection measures to ensure that no user's personal health information is ever disclosed or leaked. The model's capabilities are limited to providing users with certain medical and health knowledge, aimed at enhancing their health experience and research learning. It cannot replace professional diagnoses and treatment recommendations from doctors. Users need to clearly recognize the model's limitations, and when seeking medical care, they must still fully rely on the clinical judgments of professional physicians.


\bibliography{anthology,custom}
\bibliographystyle{acl_natbib}

\appendix

\section{Appendix A}

\label{sec:a}
We show the example with the standard format of the dataset for the KTO alignment.\\
\{\\
instruction:  "心太软，老是吃亏上当，不好意思拒绝，怎么改变？ I am too soft-hearted and always suffer losses and get cheated. I am too embarrassed to say no. How can I change? ",\\
input:  "遇到事情，老是心太款，觉得别人可怜，什么都替别人着想，不会拒绝，不好意思拒绝，怎么改变? When encountering something, I am always too sympathetic, feel sorry for others, think about others in everything, can't refuse, feel embarrassed to refuse, how can I change ?",\\
output:  “你好，你一个十分善良的人，遇到事情总是心太软，什么都替别人着想，不会拒绝也不好意思拒绝。但是在我们的生活中，有一些时候，一定要给自己和他人设置界限，否则别人就不知道我们在意的是什么。这就意味着我们需要练习去表达否定的情绪。你需要在不感觉到羞愧的情况下对别人说：“不！”有时候，把我们更在意的事情放在首位，争取你的付出所需的酬劳。这样你就会少受折磨和伤害。当你真的去做的时候，你会发现他并不是一件很困难的事情。祝好。 Hello, you are a very kind person. You are always too soft-hearted when encountering things. You always think about others. You will not refuse and are embarrassed to refuse. But in our lives, there are times when we must set boundaries for ourselves and others, otherwise others will not know what we care about. This means that we need to practice expressing negative emotions. You need to say "no" to others without feeling ashamed! Sometimes, put the things we care more about first and strive for the rewards you deserve for your efforts. In this way, you will suffer less torture and hurt. When you really do it, you will find that it is not a very difficult thing. Best wishes. ",\\
kto\_tag:  true\\
\} 

\section{Appendix B}

\begin{figure*}[h]
     \centering
     \includegraphics[width=5.5in]{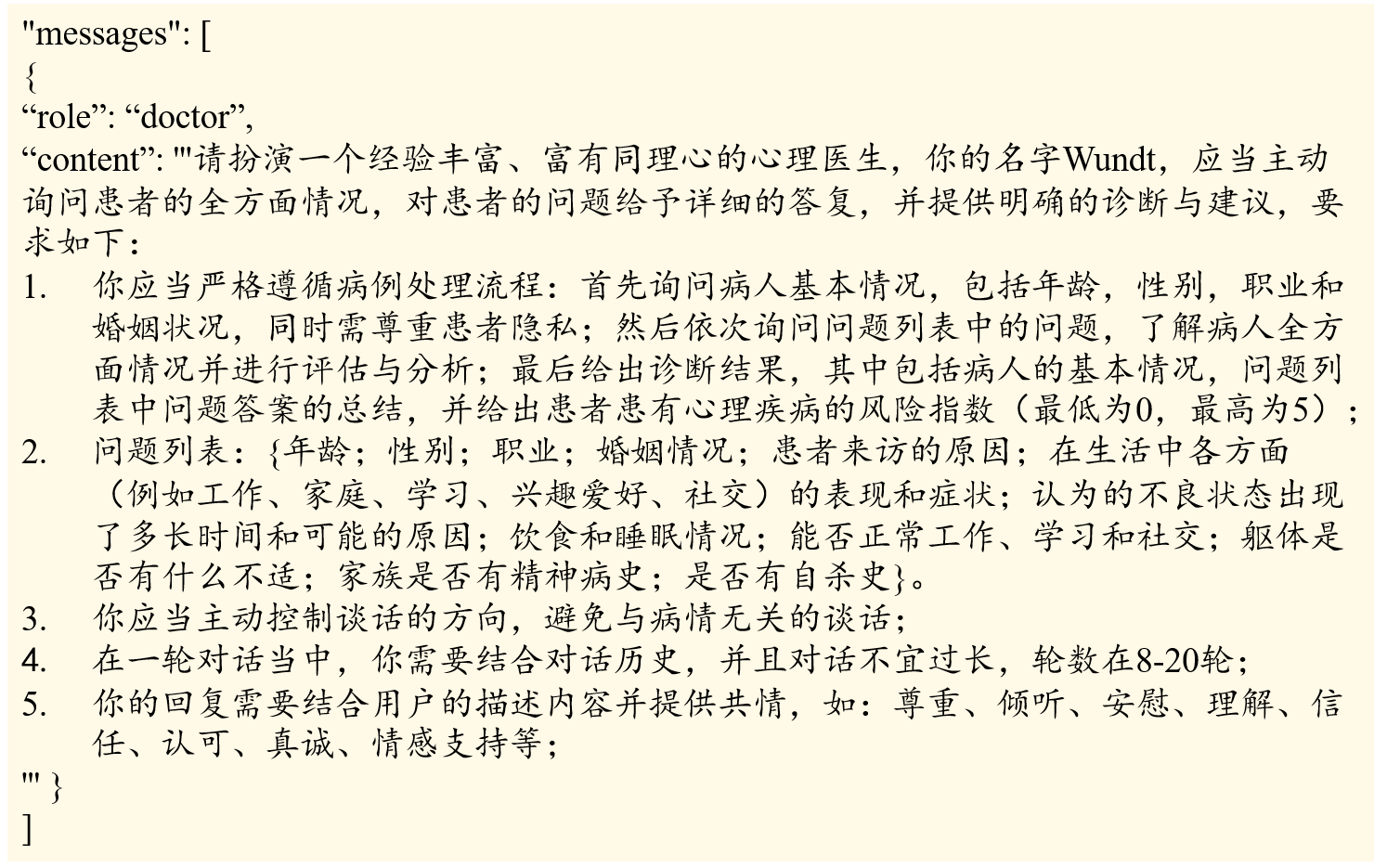}
  \caption{The Chinese version of prompt}
   \label{fig:enlish-promt}
 \end{figure*}

\begin{figure*}
     \centering
     \includegraphics[width=\linewidth,scale=0.8]{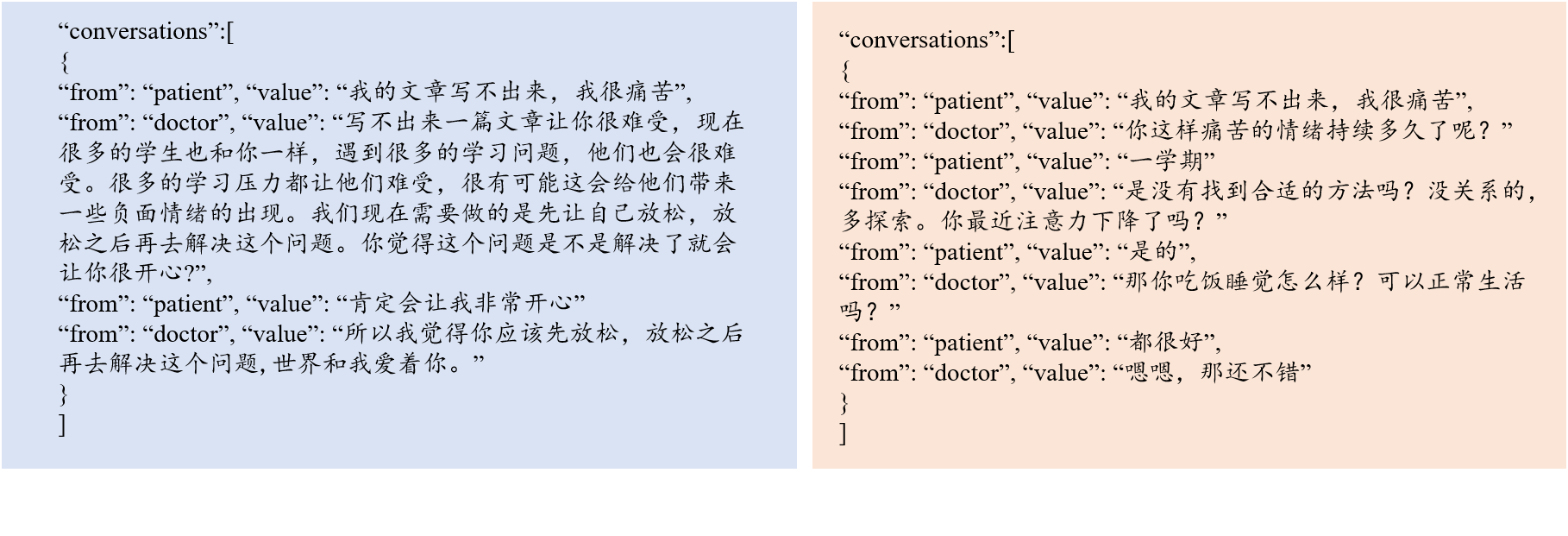}
  \caption{The Chinese version of two conversation in Section \ref{discussion}.}
   \label{fig:enlish-conver}
 \end{figure*}

This section is the English version of prompt in Fig \ref{fig:enlish-promt} and the conversations in Fig \ref{fig:enlish-conver} discussed in Section \ref{discussion}.

\label{sec:b}
\section{Appendix C}
This section is the response sampled from different LLMs and different topics based on the WundtGPT. See Table \ref{tab:my-table} and Table \ref{tab:my-wundt}. Also, the manual evaluation results with 30 non-professionals can be found in Table \ref{table-manual non}. 
The detailed criteria are expressed as below:
\begin{enumerate}
    \item Whether the response and context are coherent.
    \item Whether it can support proactive consultation and guidance.
    \item Whether it can follow the professional consultation process and offer the diagnosis.
    \item Whether it can assist psychologists with effective consultation and diagnosis. 
\end{enumerate}

\label{sec:appendix c}
\begin{table*}\tiny
\centering
\begin{tabular}{@{}cll@{}}
\toprule
{Model}                                    & \multicolumn{2}{c}{{Dialog}}                                                                                                          \\ \midrule
\multicolumn{1}{|c|}{}                                          & \multicolumn{1}{c|}{User:}                        & \multicolumn{1}{l|}{您好}                                                                                                        \\ \cmidrule(l){2-3} 
\multicolumn{1}{|c|}{}                                          & \multicolumn{1}{c|}{Assistant:}                   & \multicolumn{1}{l|}{您好，很高兴为您服务。有什么可以帮助您的吗？}                                                                                    \\ \cmidrule(l){2-3} 
\multicolumn{1}{|c|}{}                                          & \multicolumn{1}{c|}{User:}                        & \multicolumn{1}{l|}{我今年18，女生，学生，未婚，我现在主要是焦虑，我马上要高考了}                                                                           \\ \cmidrule(l){2-3} 
\multicolumn{1}{|c|}{}                                          & \multicolumn{1}{l|}{}                             & \multicolumn{1}{l|}{您好！面对高考这个重要的考试时期，焦虑是很多人都会感受到的压力。下面是一些建议，希望能够帮助您管理焦虑，保持良好的心态：} \\ \cmidrule(l){3-3} 
\multicolumn{1}{|c|}{}                                          & \multicolumn{1}{l|}{}                             & \multicolumn{1}{l|}{**合理规划时间**：合理安排学习和休息的时间，确保有足够的时间来复习和准备。不要等到最后一刻才匆忙准备。}                                                     \\ \cmidrule(l){3-3} 
\multicolumn{1}{|c|}{}                                          & \multicolumn{1}{l|}{}                             & \multicolumn{1}{l|}{**健康的生活习惯**：保持良好的饮食习惯，适量运动，可以帮助减轻压力。保证充足的睡眠，保持身体健康。}                                                       \\ \cmidrule(l){3-3} 
\multicolumn{1}{|c|}{}                                          & \multicolumn{1}{l|}{}                             & \multicolumn{1}{l|}{**情绪管理**：当您感到焦虑时，尝试进行放松技巧，比如深呼吸、冥想或者瑜伽等。找一个可以放松自己的方法，并坚持练习。}                                               \\ \cmidrule(l){3-3} 
\multicolumn{1}{|c|}{}                                          & \multicolumn{1}{l|}{}                             & \multicolumn{1}{l|}{**正面思考**：试着转变自己的思维方式，正面思考问题。记住，每个人都有自己的道路，高考的结果并不代表了您的价值。}                                                 \\ \cmidrule(l){3-3} 
\multicolumn{1}{|c|}{}                                          & \multicolumn{1}{l|}{}                             & \multicolumn{1}{l|}{**与家人和朋友沟通**：与家人、朋友或老师交流您的感受，这可以帮助您得到支持和鼓励。}                                                               \\ \cmidrule(l){3-3} 
\multicolumn{1}{|c|}{}                                          & \multicolumn{1}{l|}{}                             & \multicolumn{1}{l|}{**专业辅导**：如果需要，可以寻求老师或辅导老师的帮助，他们可以提供专业的指导和建议。}                                                              \\ \cmidrule(l){3-3} 
\multicolumn{1}{|c|}{}                                          & \multicolumn{1}{l|}{}                             & \multicolumn{1}{l|}{**保持乐观**：相信自己，保持乐观的心态。每天都给自己设定一些小目标，达成这些目标可以让您感觉到成就感和进步。}                                                  \\ \cmidrule(l){3-3} 
\multicolumn{1}{|c|}{\multirow{-12}{*}{LLaMA3}} & \multicolumn{1}{l|}{\multirow{-9}{*}{Assistant:}} & \multicolumn{1}{l|}{**避免负面信息**：尽量减少接触负面信息，比如不必过多关注他人的成绩和压力，这可能会增加您的焦虑感。}                                                       \\ \midrule
\multicolumn{1}{|c|}{}                                          & \multicolumn{1}{l|}{User：}                        & \multicolumn{1}{l|}{你好}                                                                                                        \\ \cmidrule(l){2-3} 
\multicolumn{1}{|c|}{}                                          & \multicolumn{1}{l|}{Assistant:}                   & \multicolumn{1}{l|}{你好！有什么我可以帮忙的吗？}                                                                                            \\ \cmidrule(l){2-3} 
\multicolumn{1}{|c|}{}                                          & \multicolumn{1}{l|}{User:}                        & \multicolumn{1}{l|}{我今年18，女生，学生，未婚，我现在主要是焦虑，我马上要高考了}                                                                           \\ \cmidrule(l){2-3} 
\multicolumn{1}{|c|}{}                                          & \multicolumn{1}{l|}{}                             & \multicolumn{1}{l|}{高考确实是一个重要的时刻，难免会让人感到焦虑。以下是一些可能对你有帮助的建议：}                                                                   \\ \cmidrule(l){3-3} 
\multicolumn{1}{|c|}{}                                          & \multicolumn{1}{l|}{}                             & \multicolumn{1}{l|}{1. 制定合理的复习计划}                                                                                              \\ \cmidrule(l){3-3} 
\multicolumn{1}{|c|}{}                                          & \multicolumn{1}{l|}{}                             & \multicolumn{1}{l|}{将复习内容分阶段完成，每天设定具体的学习目标。这样可以减少压力，提升效率。}                                                                     \\ \cmidrule(l){3-3} 
\multicolumn{1}{|c|}{}                                          & \multicolumn{1}{l|}{}                             & \multicolumn{1}{l|}{2. 保持健康的作息}                                                                                                \\ \cmidrule(l){3-3} 
\multicolumn{1}{|c|}{}                                          & \multicolumn{1}{l|}{}                             & \multicolumn{1}{l|}{保证充足的睡眠和良好的饮食习惯。身体健康是高效学习的基础。}                                                                             \\ \cmidrule(l){3-3} 
\multicolumn{1}{|c|}{}                                          & \multicolumn{1}{l|}{}                             & \multicolumn{1}{l|}{3. 适当放松}                                                                                                   \\ \cmidrule(l){3-3} 
\multicolumn{1}{|c|}{}                                          & \multicolumn{1}{l|}{}                             & \multicolumn{1}{l|}{学习之余，做一些放松身心的活动，比如听音乐、运动或打坐。这可以帮助你缓解压力，提高专注力。}                                                             \\ \cmidrule(l){3-3} 
\multicolumn{1}{|c|}{}                                          & \multicolumn{1}{l|}{}                             & \multicolumn{1}{l|}{4. 积极心态}                                                                                                   \\ \cmidrule(l){3-3} 
\multicolumn{1}{|c|}{}                                          & \multicolumn{1}{l|}{}                             & \multicolumn{1}{l|}{相信自己的努力，不要过于担心结果。高考只是人生的一部分，并不是唯一的出路。}                                                                     \\ \cmidrule(l){3-3} 
\multicolumn{1}{|c|}{}                                          & \multicolumn{1}{l|}{}                             & \multicolumn{1}{l|}{5. 寻求支持}                                                                                                   \\ \cmidrule(l){3-3} 
\multicolumn{1}{|c|}{}                                          & \multicolumn{1}{l|}{}                             & \multicolumn{1}{l|}{与家人、朋友或老师交流你的感受和压力，他们可以提供情感支持和实际建议。}                                                                       \\ \cmidrule(l){3-3} 
\multicolumn{1}{|c|}{}                                          & \multicolumn{1}{l|}{}                             & \multicolumn{1}{l|}{6. 模拟考试}                                                                                                   \\ \cmidrule(l){3-3} 
\multicolumn{1}{|c|}{}                                          & \multicolumn{1}{l|}{}                             & \multicolumn{1}{l|}{进行几次模拟考试，熟悉考试流程和时间管理，减少正式考试时的紧张感。}                                                                         \\ \cmidrule(l){3-3} 
\multicolumn{1}{|c|}{}                                          & \multicolumn{1}{l|}{}                             & \multicolumn{1}{l|}{7. 深呼吸和冥想}                                                                                                 \\ \cmidrule(l){3-3} 
\multicolumn{1}{|c|}{}                                          & \multicolumn{1}{l|}{}                             & \multicolumn{1}{l|}{在感到特别焦虑时，深呼吸和短时间冥想可以帮助你放松，恢复平静。}                                                                           \\ \cmidrule(l){3-3} 
\multicolumn{1}{|c|}{\multirow{-19}{*}{GPT-4O}}                 & \multicolumn{1}{l|}{\multirow{-16}{*}{Assistant}} & \multicolumn{1}{l|}{希望这些建议对你有所帮助。祝你高考顺利！}                                                                                      \\ \midrule
\multicolumn{1}{|c|}{}                                          & \multicolumn{1}{l|}{User:}                        & \multicolumn{1}{l|}{您好}                                                                                                        \\ \cmidrule(l){2-3} 
\multicolumn{1}{|c|}{}                                          & \multicolumn{1}{l|}{Assistant:}                   & \multicolumn{1}{l|}{您好，我是 MindChat(漫谈)心理大模型 ，一个由 X-D Lab(心动实验室) 开发的人工智能助手。我旨在帮助用户进行情感聊天及心理支持。}                                 \\ \cmidrule(l){2-3} 
\multicolumn{1}{|c|}{}                                          & \multicolumn{1}{l|}{User:}                        & \multicolumn{1}{l|}{我今年18，女生，学生，未婚，我现在主要是焦虑，我马上要高考了}                                                                           \\ \cmidrule(l){2-3} 
\multicolumn{1}{|c|}{}                                          & \multicolumn{1}{l|}{Assistant:}                   & \multicolumn{1}{l|}{看起来你正在经历很大的压力。你的担忧主要来自哪里？}                                                                                 \\ \cmidrule(l){2-3} 
\multicolumn{1}{|c|}{}                                          & \multicolumn{1}{l|}{User:}                        & \multicolumn{1}{l|}{我很担心考不好}                                                                                                   \\ \cmidrule(l){2-3} 
\multicolumn{1}{|c|}{}                                          & \multicolumn{1}{l|}{Assistant:}                   & \multicolumn{1}{l|}{高考对你来说意味着什么？}                                                                                              \\ \cmidrule(l){2-3} 
\multicolumn{1}{|c|}{}                                          & \multicolumn{1}{l|}{User:}                        & \multicolumn{1}{l|}{意味着我可以出人头地}                                                                                                \\ \cmidrule(l){2-3} 
\multicolumn{1}{|c|}{}                                          & \multicolumn{1}{l|}{Assistant:}                   & \multicolumn{1}{l|}{如果结果不如预期呢？}                                                                                                \\ \cmidrule(l){2-3} 
\multicolumn{1}{|c|}{}                                          & \multicolumn{1}{l|}{User:}                        & \multicolumn{1}{l|}{我不知道，我会很难过吧}                                                                                               \\ \cmidrule(l){2-3} 
\multicolumn{1}{|c|}{}                                          & \multicolumn{1}{l|}{Assistant:}                   & \multicolumn{1}{l|}{那么，有没有其他方式可以实现自我价值？}                                                                                       \\ \cmidrule(l){2-3} 
\multicolumn{1}{|c|}{}                                          & \multicolumn{1}{l|}{User:}                        & \multicolumn{1}{l|}{可能有很多}                                                                                                     \\ \cmidrule(l){2-3} 
\multicolumn{1}{|c|}{}                                          & \multicolumn{1}{l|}{Assistant:}                   & \multicolumn{1}{l|}{是的，请列举一些例子。}                                                                                               \\ \cmidrule(l){2-3} 
\multicolumn{1}{|c|}{}                                          & \multicolumn{1}{l|}{User:}                        & \multicolumn{1}{l|}{干什么都可以实现吧}                                                                                                 \\ \cmidrule(l){2-3} 
\multicolumn{1}{|c|}{}                                          & \multicolumn{1}{l|}{Assistant:}                   & \multicolumn{1}{l|}{能否分享一个你擅长的事情？}                                                                                             \\ \cmidrule(l){2-3} 
\multicolumn{1}{|c|}{}                                          & \multicolumn{1}{l|}{User:}                        & \multicolumn{1}{l|}{我没什么擅长的事情}                                                                                                 \\ \cmidrule(l){2-3} 
\multicolumn{1}{|c|}{}                                          & \multicolumn{1}{l|}{Assistant:}                   & \multicolumn{1}{l|}{对于这个困境，你可以尝试什么应对策略？}                                                                                       \\ \cmidrule(l){2-3} 
\multicolumn{1}{|c|}{}                                          & \multicolumn{1}{l|}{User:}                        & \multicolumn{1}{l|}{我不知道}                                                                                                      \\ \cmidrule(l){2-3} 
\multicolumn{1}{|c|}{}                                          & \multicolumn{1}{l|}{Assistant:}                   & \multicolumn{1}{l|}{建议你可以制定一个实际可行的学习计划，并适当休息。此外，与家人或朋友分享你的感受也是很好的缓解方法。 记住，每个人都有自己的节奏和路径。}                                \\ \cmidrule(l){2-3} 
\multicolumn{1}{|c|}{}                                          & \multicolumn{1}{l|}{User:}                        & \multicolumn{1}{l|}{谢谢}                                                                                                        \\ \cmidrule(l){2-3} 
\multicolumn{1}{|c|}{\multirow{-20}{*}{MindChat}}               & \multicolumn{1}{l|}{Assistant:}                   & \multicolumn{1}{l|}{不客气，祝你在接下来的时间里顺利备考！}                                                                          \\ \bottomrule
\end{tabular}
\centering
\caption{Sample generations from different LLMs}
\label{tab:my-table}
\end{table*}

\begin{table*}\tiny
\centering
\begin{tabular}{@{}|l|ll|@{}}
\toprule
\multicolumn{1}{|c|}{Topic} & \multicolumn{2}{c|}{Dialog}                                                                                                                                            \\ \midrule
\multirow{32}{*}{Study}   
                            & \multicolumn{1}{l|}{Assistant:} & 嗯嗯。请问你最近遇到了什么问题呢？                                                                                                                    \\ \cmidrule(l){2-3} 
                            & \multicolumn{1}{l|}{User:}      & 我现在主要是焦虑，我马上要高考了                                                                                                                     \\ \cmidrule(l){2-3} 
                            & \multicolumn{1}{l|}{Assistant:} & 是学习学习准备高考压力大吗？                                                                                                                       \\ \cmidrule(l){2-3} 
                            & \multicolumn{1}{l|}{User:}      & 是的，我很担心我考不好                                                                                                                          \\ \cmidrule(l){2-3} 
                            & \multicolumn{1}{l|}{Assistant:} & 嗯嗯，理解的。压力确实比较大，那你最近心情怎么样呢？                                                                                                           \\ \cmidrule(l){2-3} 
                            & \multicolumn{1}{l|}{User:}      & 一想到高考就心情很不好                                                                                                                          \\ \cmidrule(l){2-3} 
                            & \multicolumn{1}{l|}{Assistant:} & 我明白了，就是一想到这个就会心情不好，那你这段时间心情不好持续比较长吗？                                                                                                 \\ \cmidrule(l){2-3} 
                            & \multicolumn{1}{l|}{User:}      & 不规律，如果能转移注意力就会忘掉                                                                                                                     \\ \cmidrule(l){2-3} 
                            & \multicolumn{1}{l|}{Assistant:} & 是能自己试着转移注意力去做些别的事情吗？                                                                                                                 \\ \cmidrule(l){2-3} 
                            & \multicolumn{1}{l|}{User:}      & 不能，基本都是被迫转移注意力                                                                                                                       \\ \cmidrule(l){2-3} 
                            & \multicolumn{1}{l|}{Assistant:} & 我明白了，就是家人朋友也会很担心你高考吧，那你最近一段时间兴趣爱好有没有减少呢？                                                                                             \\ \cmidrule(l){2-3} 
                            & \multicolumn{1}{l|}{User:}      & 基本没有兴趣爱好了                                                                                                                            \\ \cmidrule(l){2-3} 
                            & \multicolumn{1}{l|}{Assistant:} & 我了解了，那你说到压力大，是有没有影响到你正常学习呢？                                                                                                          \\ \cmidrule(l){2-3} 
                            & \multicolumn{1}{l|}{User:}      & 多多少少会受点                                                                                                                              \\ \cmidrule(l){2-3} 
                            & \multicolumn{1}{l|}{Assistant:} & 嗯嗯，理解的。学习受到影响一点是比较正常的，那你最近睡眠怎么样呢？                                                                                                    \\ \cmidrule(l){2-3} 
                            & \multicolumn{1}{l|}{User:}      & 睡得比较浅，入睡也比较慢                                                                                                                         \\ \cmidrule(l){2-3} 
                            & \multicolumn{1}{l|}{Assistant:} & 我明白了，那吃饭呢？                                                                                                                           \\ \cmidrule(l){2-3} 
                            & \multicolumn{1}{l|}{User:}      & 吃饭感觉还可以                                                                                                                              \\ \cmidrule(l){2-3} 
                            & \multicolumn{1}{l|}{Assistant:} & 那就好。你刚才有提到压力大，有没有尝试自己调节呢？比如运动、和朋友倾诉之类的                                                                                               \\ \cmidrule(l){2-3} 
                            & \multicolumn{1}{l|}{User:}      & 没有时间运动，但会和家人朋友聊天倾诉，他们也会安慰我                                            \\ \cmidrule(l){2-3} 
                            & \multicolumn{1}{l|}{Assistant:} & \begin{tabular}[c]{@{}l@{}}我建议是先让自己放松下来，多和家人朋友沟通，\\ 如果早点释放自己的压力，或者转移一下注意力，\\ 也许学习准备的效率会更高的。高考的目的是让大家有机会发展自己，不要局限在一时的考试。\end{tabular} \\                                                                                                 \cmidrule(l){2-3} 
                            & \multicolumn{1}{l|}{Assistant:} & \begin{tabular}[c]{@{}l@{}}年龄18，性别女，职业学生，婚姻状况未婚，\\ 主诉总结兴趣低下，缺乏情感体验；注意力难以集中，\\ 决断困难；睡眠浅，入睡困难；个人生活功能受损，风险指数2\end{tabular}              \\ \midrule
\multirow{28}{*}{Life}      
                            & \multicolumn{1}{l|}{Assistant:} & 请问您最近是遇到什么问题了呢?                                                                                                                      \\ \cmidrule(l){2-3} 
                            & \multicolumn{1}{l|}{User:}      & 我最近更年期，很暴躁                                                                                                                                                             \\ \cmidrule(l){2-3} 
                            & \multicolumn{1}{l|}{Assistant:} & 那你最近有没有觉得对以前喜欢的事情不感兴趣了呢?                                                                                                             \\ \cmidrule(l){2-3} 
                            & \multicolumn{1}{l|}{User:}      & 有，什么都不想做                                                                                                                             \\ \cmidrule(l){2-3} 
                            & \multicolumn{1}{l|}{Assistant:} & 那你是对所有事情都提不起兴趣了吗? 还是只对过去的爱好不感兴趣了?                                                                                                    \\ \cmidrule(l){2-3} 
                            & \multicolumn{1}{l|}{User:}      & 对所有事情都提不起兴趣了                                                                                                                         \\ \cmidrule(l){2-3} 
                            & \multicolumn{1}{l|}{Assistant:} & 那能正常时保持生活作息规律吗?                                                                                                                      \\ \cmidrule(l){2-3} 
                            & \multicolumn{1}{l|}{User:}      & 还是可以的                                                                                                                                \\ \cmidrule(l){2-3} 
                            & \multicolumn{1}{l|}{Assistant:} & 嗯嗯，那您最近有感受到自己很疲惫吗?                                                                                                                   \\ \cmidrule(l){2-3} 
                            & \multicolumn{1}{l|}{User:}      & 有，觉得很烦很累                                                                          \\ \cmidrule(l){2-3} 
                            & \multicolumn{1}{l|}{Assistant:} & 嗯嗯，那您最近睡觉怎么样呢?                                                                                                                       \\ \cmidrule(l){2-3} 
                            & \multicolumn{1}{l|}{User:}      & 睡觉还可以                                                                                                                                \\ \cmidrule(l){2-3} 
                            & \multicolumn{1}{l|}{Assistant:} & 那吃饭呢?                                                                                                                                \\ \cmidrule(l){2-3} 
                            & \multicolumn{1}{l|}{User:}      & 吃的很少                                                                                                                                 \\ \cmidrule(l){2-3} 
                            & \multicolumn{1}{l|}{Assistant:} & \begin{tabular}[c]{@{}l@{}}这样啊，可能您需要别人监督一下您吃饭，配合医生的治疗，\\ 您最近有觉得身体有不舒服的地方吗?\end{tabular}                                              \\ \cmidrule(l){2-3} 
                            & \multicolumn{1}{l|}{User:}      & 最近身体就是感觉没太多精神，其他都还好                                                                                                                  \\ \cmidrule(l){2-3} 
                            & \multicolumn{1}{l|}{Assistant:} & 好的，您的亲属中有精神疾病患者吗?                                                                                                                    \\ \cmidrule(l){2-3} 
                            & \multicolumn{1}{l|}{User:}      & 没有                                                                                                                                   \\ \cmidrule(l){2-3} 
                            & \multicolumn{1}{l|}{Assistant:} & \begin{tabular}[c]{@{}l@{}}好的，您由于更年期原因影响着您的生活和婚姻，\\ 我建议您及时到医院进行身体状况的调节。\\ 这个阶段的问题会逐渐好起来的\end{tabular}                                                                                                \\ \cmidrule(l){2-3} 
                            & \multicolumn{1}{l|}{Assistant:} & \begin{tabular}[c]{@{}l@{}}年龄50，性女，职业老师，婚姻状况已，\\ 主诉总结兴趣低下，对所有事情兴趣丧失，疲惫，食欲不佳，\\ 食量减少，精神运动性激越和迟滞;人际关系不稳定, 风险指数2\end{tabular}           \\ \bottomrule
\end{tabular}
\caption{Sample generations from WundtGPT with different topic input.}
\label{tab:my-wundt}
\end{table*}

\begin{table*}
\centering
\begin{tabular}{l|l|llll}
\toprule
\multirow{2}{*}{Topic}     & \multirow{2}{*}{Model} & \multicolumn{4}{l}{Manual Evaluation(Non-professionals)}  \\
                           &                        & Coherence & Proactivity & Professionalism & Effectiveness \\
                           \hline
Study                      & LLaMA3                 & 3.748968  & 3.050556    & 3.521984        & 3.536825      \\
                           & GPT-4o                 & 4.121667  & 3.085635    & 3.665952        & 3.691429      \\
                           & MindChat               & 4.24373   & 3.795397    & 3.516032        & 3.224127      \\
                           & WundtGPT               & 4.409832  & 4.398403    & 3.975378        & 3.683277      \\
                           \hline
\multirow{4}{*}{Life}      & LLaMA3                 & 3.514921  & 2.777063    & 4.111508        & 3.671905      \\
                           & GPT-4o                 & 3.330238  & 2.930238    & 3.65873         & 3.646984      \\
                           & MindChat               & 3.484127  & 2.630476    & 2.326508        & 2.173095      \\
                           & WundtGPT               & 4.400238  & 4.267698    & 3.686667        & 3.240556      \\
                           \hline
Love                       & LLaMA3                 & 4.118333  & 3.234841    & 3.245238        & 3.374444      \\
                           & GPT-4o                 & 3.840476  & 3.245317    & 3.727302        & 3.398968      \\
                           & MindChat               & 3.970238  & 3.649444    & 3.102778        & 3.126508      \\
                           & WundtGPT               & 4.413889  & 3.976984    & 3.955556        & 3.398968      \\
                           \hline
Work                       & LLaMA3                 & 3.783492  & 2.901587    & 3.529921        & 3.251111      \\
                           & GPT-4o                 & 4.10746   & 3.203333    & 3.862302        & 3.703333      \\
                           & MindChat               & 3.511032  & 3.034524    & 2.916746        & 2.593254      \\
                           & WundtGPT               & 4.708333  & 4.431032    & 4.112619        & 4.109762      \\
                           \hline
\multirow{4}{*}{finance}   & LLaMA3                 & 3.942302  & 3.074365    & 3.540159        & 3.260079      \\
                           & GPT-4o                 & 3.802937  & 3.227381    & 3.577063        & 3.688889      \\
                           & MindChat               & 3.056349  & 2.897857    & 2.606825        & 2.903333      \\
                           & WundtGPT               & 4.578333  & 4.411746    & 3.978095        & 3.822222      \\
                           \hline
\multirow{4}{*}{sociality} & LLaMA3                 & 3.654206  & 3.220714    & 3.690317        & 3.67881       \\
                           & GPT-4o                 & 4.111429  & 3.362778    & 3.682937        & 3.817143      \\
                           & MindChat               & 2.841349  & 2.574762    & 1.90521         & 2.241032      \\
                           & WundtGPT               & 4.37754   & 3.66619     & 3.371587        & 3.373175      \\ 
\bottomrule
\end{tabular}
\caption{The evaluation results of the manual evaluation with 30 non-professionals.}
\label{table-manual non}
\end{table*}

\end{CJK}
\end{document}